\documentclass{article}

\usepackage{arxiv}

\usepackage[utf8]{inputenc} 
\usepackage[T1]{fontenc}    
\usepackage{hyperref}       
\usepackage{url}            
\usepackage{booktabs}       
\usepackage{amsfonts}       
\usepackage{nicefrac}       
\usepackage{microtype}      
\usepackage{lipsum}		
\usepackage{graphicx}
\usepackage[numbers, sort&compress]{natbib}
\usepackage{doi}
\usepackage{comment}
\usepackage[table]{xcolor}
\usepackage{amsmath}
\usepackage{multirow}
\usepackage{wrapfig}
\usepackage{caption}

\title{Routing without Forgetting}

\author{{\href{https://orcid.org/0009-0001-1806-5887}       {\includegraphics[scale=0.06]{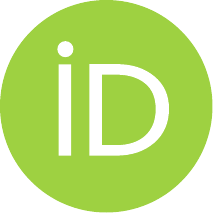}\hspace{1mm}\hspace{1mm}Alessio~Masano}}  \\
	University of Catania\\
	\texttt{alessio.masano@phd.unict.it} \\
	\And
	{\href{https://orcid.org/0000-0002-1333-8348}{\includegraphics[scale=0.06]{orcid.pdf}\hspace{1mm}\hspace{1mm}Giovanni~Bellitto}} \\
	University of Catania\\
	\texttt{giovanni.bellitto@unict.it} \\
	\AND
    {\href{https://orcid.org/0000-0002-3856-603X}{\includegraphics[scale=0.06]{orcid.pdf}\hspace{1mm}\hspace{1mm}Dipam~Goswami}} \\
    Department of Computer Science, Universitat Autònoma de Barcelona \\
	Computer Vision Center, Barcelona\\
	\texttt{dgoswami@cvc.uab.cat} \\
    \And
    {\href{https://orcid.org/0000-0002-9656-9706}{\includegraphics[scale=0.06]{orcid.pdf}\hspace{1mm}\hspace{1mm}Joost Van de Weijer}} \\
    Department of Computer Science, Universitat Autònoma de Barcelona \\
	Computer Vision Center, Barcelona\\
	\texttt{joost@cvc.uab.es} \\
    \AND
    {\href{https://orcid.org/0000-0001-6653-2577}{\includegraphics[scale=0.06]{orcid.pdf}\hspace{1mm}\hspace{1mm}Concetto~Spampinato}} \\
	University of Catania\\
	\texttt{concetto.spampinato@unict.it} \\
}

\date{}

\renewcommand{\shorttitle}{Routing without Forgetting}

\hypersetup{
pdftitle={A template for the arxiv style},
pdfsubject={q-bio.NC, q-bio.QM},
pdfauthor={David S.~Hippocampus, Elias D.~Striatum},
pdfkeywords={First keyword, Second keyword, More},
}

\begin{document}
\maketitle

\begin{abstract}
Continual learning in transformers is commonly addressed through parameter-efficient adaptation: prompts, adapters, or LoRA modules are specialized per task while the backbone remains frozen. Although effective in controlled multi-epoch settings, these approaches rely on gradual gradient-based specialization and struggle in \emph{Online Continual Learning (OCL)}, where data arrive as a non-stationary stream and each sample may be observed only once.

We recast continual learning in transformers as a routing problem: under strict online constraints, the model must dynamically select the appropriate representational subspace for each input without explicit task identifiers or repeated optimization. We thus introduce \emph{Routing without Forgetting (RwF)}, a transformer architecture augmented with energy-based associative retrieval layers inspired by Modern Hopfield Networks. Instead of storing or merging task-specific prompts, RwF generates dynamic prompts through single-step associative retrieval over the transformer token embeddings at each layer. Retrieval corresponds to the closed-form minimization of a strictly convex free-energy functional, enabling input-conditioned routing within each forward pass, independently of iterative gradient refinement.

Across challenging class-incremental benchmarks, RwF improves over existing prompt-based methods. On Split-ImageNet-R and Split-ImageNet-S, RwF outperforms prior prompt-based approaches by a large margin, even in few-shot learning regimes. These results indicate that embedding energy-based associative routing directly within the transformer backbone provides a principled and effective foundation for OCL.
\end{abstract}


\section{Introduction}
\label{sec:intro}
As pre-trained Vision Transformers (ViTs)~\cite{vit} have become the dominant backbone for Continual Learning (CL), adaptation has shifted from updating all network weights to managing small, task-conditioned parameter modules. Recent transformer-based approaches adopt parameter-efficient strategies, such as prompt pools~\cite{l2p,dualprompt,coda}, adapters~\cite{aper,ease}, or LoRA~\cite{hu2022lora,inflora} modules, that specialize subsets of parameters while keeping the backbone largely frozen. These methods have demonstrated strong performance in class-incremental benchmarks. However, stability is typically achieved through the gradual specialization, selection, or composition of task-associated parameter subsets via iterative gradient updates. In \emph{Online Continual Learning} (OCL), where data arrive as a non-stationary stream and may be observed only once, adaptation mechanisms that depend primarily on repeated optimization can become reactive: routing decisions must be progressively corrected through successive parameter updates.

We argue that CL in transformer architectures can instead be framed as a routing problem. Under strict online constraints, the model should dynamically select the appropriate representational subspace for each input without explicit task identifiers and without relying solely on accumulated parameter specialization. Rather than allocating or merging task-indexed modules, we propose to modify how representations are selected and combined within the backbone itself. Thus, in this paper we propose \textbf{Routing without Forgetting (RwF)}, a transformer architecture augmented with energy-based associative retrieval layers inspired by Modern Hopfield Networks~\cite{ramsauer2021hopfield}. Instead of relying on task-conditioned parameter modules, RwF generates input-conditioned routing prompts through single-step associative retrieval over the current feature sequence. These prompts correspond to the minimizers of a strictly convex free-energy functional, yielding a unique equilibrium distribution in closed form. Routing decisions are therefore computed analytically within each forward pass, while parameters remain updated via gradient descent across the stream. Representation selection is thus partially decoupled from optimization dynamics. This architectural modification is particularly suited to OCL. Because routing is recomputed at every layer and every forward pass, representation selection adapts immediately to distribution changes even before parameters have converged. Moreover, the retrieval operator is continuous and input-smooth, helping mitigate abrupt representational transitions under streaming drift.\\
\indent We evaluate RwF under strict online class-incremental settings on Split-CIFAR-100, Split-ImageNet-R, and Split-ImageNet-S. On the large-scale ImageNet benchmarks, RwF consistently outperforms strong prompt-based~\cite{l2p,dualprompt,coda} and LoRA-based~\cite{onlinelora,inflora} baselines while introducing only 2.1\% additional trainable parameters, remaining within the parameter-efficient regime. 
Under the strict single-pass protocol of~\cite{onlinelora}, RwF achieves 74.09\% final average accuracy on Split-ImageNet-R and 61.37\% on Split-ImageNet-S. Beyond overall accuracy, RwF remains robust under progressively reduced supervision and increasing task fragmentation. In few-shot regimes, it maintains a clear margin as training samples per task decrease, exhibiting more controlled degradation than prompt- and adapter-based methods. As the number of sequential tasks grows (from 5 to 40), RwF preserves a consistent advantage, indicating improved scalability under frequent distribution shifts.  Controlled ablations further show that routing depth systematically modulates performance, demonstrating that RwF alters internal transformer routing dynamics rather than adding task-conditioned parameters.\\
\indent Together, these results indicate that embedding energy-based associative routing within the backbone provides an effective mechanism for online continual learning in transformers.

\section{Related work}
\label{sec:related}
We review prior work through the lens of \emph{how routing and stability are implemented} in CL. Existing approaches typically constrain parameter updates, introduce explicit selection mechanisms (e.g., experts or prompts), or rely on memory buffers. In contrast, we cast continual learning as an \emph{energy-based associative routing problem} embedded directly within a single transformer backbone.\\
\emph{Stability Mechanisms in Continual Learning.}
Most continual learning methods mitigate catastrophic forgetting by constraining optimization dynamics. Replay-based approaches interleave stored samples to reduce representational drift~\cite{agem,er,gdumb,mir,pennisi2023experience,10695036,pcr}, while regularization-based methods penalize deviations from previously important parameters~\cite{ewc,ewcpp,icarl,Sorrenti_2023_ICCV,lucir}. Knowledge distillation aligns predictions with earlier model snapshots~\cite{lwf,podnet,boschini2022class}, and dynamic architectures allocate task-conditioned modules to limit interference~\cite{foster,memo}. Across these families, stability primarily arises from modifying or restricting gradient updates. In contrast, we explore stability as a \emph{structural property} of the architecture: by viewing routing as a continuous associative mapping, representational changes induced by distribution shifts can vary smoothly with the input.\\
\emph{Online Continual Learning (OCL).}
OCL considers a strict regime where data arrive as a non-stationary stream and each sample is typically observed only once~\cite{gem,oclsurv,oclsurvey2,ser}. Unlike multi-epoch incremental protocols, OCL requires immediate adaptation under limited optimization opportunities. Replay remains a strong baseline~\cite{gss,erace}, but single-pass, non-IID conditions can make gradient-driven specialization less reliable. This setting highlights the need for mechanisms that support fast, input-conditioned routing rather than gradual parameter drift.\\
\emph{Prompting and Parameter-Efficient Adaptation.}
Pre-trained transformers have enabled parameter-efficient continual learning through prompts, adapters, and LoRA modules~\cite{vpt,l2p,dualprompt,coda,onlinelora,inflora}. These approaches specialize, retrieve, or compose subsets of parameters while keeping the backbone largely frozen. While differing in implementation—ranging from prompt pools to modular low-rank updates—routing typically occurs at the level of parameter subsets. Such strategies can be effective under incremental protocols, but adaptation remains tied to iterative gradient-driven refinement of selected modules. In contrast, RwF inserts an associative operator directly within transformer layers, allowing the sequence to query itself and modulate internal representation dynamics without allocating task-indexed modules.\\
\emph{Modern Hopfield Networks.}
Modern Hopfield networks reinterpret attention as an energy-based associative memory with exponential capacity~\cite{ramsauer2021hopfield}. Retrieval corresponds to minimizing a strictly convex free-energy functional, yielding a unique Gibbs distribution. While Hopfield layers have been explored for memory augmentation, their role in CL remains underexplored. We leverage their variational interpretation to implement stable, input-conditioned routing within transformer layers. Unlike discrete gating mechanisms, the associative operator is continuous and differentiable, supporting smooth adaptation under gradual distribution shifts.\\
  Architecturally, RwF is conceptually related to Mixture-of-Experts (MoE) models~\cite{jacobs1991adaptive}, where adaptation is achieved through input-dependent routing. MoE architectures route inputs to specialized experts via learned gating networks~\cite{ShazeerMMDLHD17,SwitchTransformers_JMLR2022Fedus}, and expert isolation can mitigate forgetting in CL~\cite{2024divide}. RwF follows the routing principle but removes discrete expert partitioning and explicit gating. Instead, routing is obtained in closed form via energy-based associative retrieval embedded within a single transformer backbone. This yields a continuous, input-conditioned modulation of representation flow with a unique equilibrium state for fixed parameters. In online continual learning, where data are streamed and rarely revisited, this provides a structurally grounded alternative for balancing plasticity and stability without architectural growth or explicit task identifiers.

\section{Routing without Forgetting}
\label{sec:method}
\subsection{Problem Setting}
We consider the Online Continual Learning (OCL)~\cite{gss,chaudhry2021using, oclsurvey2, oclsurv,ser,onlinelora} setting, where a model $\mathcal{F}_\theta$ with parameters $\theta$ is trained on a sequence of $T$ tasks $\{\mathcal{T}_1, \mathcal{T}_2, \dots, \mathcal{T}_T\}$. Each task $\mathcal{T}_t$ is associated with a data distribution $\mathcal{D}_t$ over input-label pairs $(x, y) \in \mathcal{X} \times \mathcal{Y}$. At each step $t$, the model observes a mini-batch $\mathcal{B}_t \sim \mathcal{D}_t$ \textbf{only once}, without the possibility of revisiting past samples. 
The goal is to learn parameters $\theta$ that minimize the cumulative loss across all tasks observed so far:
\begin{equation}
    \mathcal{L}(\theta) = \sum_{i=1}^{t} \mathbb{E}_{(x,y) \sim \mathcal{D}_i} 
    \left[ \ell\left(\mathcal{F}_\theta(x), y\right) \right],
\end{equation}
while respecting the constraint that each sample is processed at most once. We follow the \textbf{Class-Incremental} (Class-IL) protocol, where the task identity is not provided at inference time and the model must discriminate among all classes. \\
This setting exposes a fundamental limitation of approaches that rely on iterative, gradient-driven specialization: when each sample—or mini-batch—is observed only once, there is insufficient opportunity for task-specific parameters to converge before the distribution shifts. Effective adaptation in OCL therefore requires mechanisms that can reconfigure internal representations \emph{immediately}, within a single forward pass, rather than over multiple optimization steps.

\subsection{Conceptual Framework}

The core philosophy of \textbf{RwF} is to move away from the ``storage'' paradigm of continual learning—where task-specific knowledge is encoded in static prompts—and toward a dynamic, input-conditioned routing mechanism embedded directly within transformer layers.

In traditional prompt tuning, adaptation is static: fixed tokens are learned through repeated gradient updates and reused at inference. In strict OCL such iterative specialization provides limited plasticity.

RwF instead introduces \emph{RwF layers}, in which each modified transformer block performs an input-conditioned associative retrieval step before self-attention. Rather than storing task-specific parameters, the model dynamically reallocates representational subspaces within the backbone at every forward pass.

This framework rests on two structural principles:

\begin{enumerate}
    \item \textbf{Energy-based retrieval:} routing corresponds to the minimization of a strictly convex free-energy functional, yielding a unique equilibrium distribution over tokens for fixed parameters.
    \item \textbf{Architectural smoothness:} routing weights depend continuously on input features, preventing abrupt representational transitions under streaming distribution shifts.
\end{enumerate}

Together, these principles allow representation reconfiguration to occur immediately in feature space, without waiting for gradient-based parameter adaptation. In OCL settings, this decouples routing speed from optimization, enabling stable adaptation under limited supervision and frequent distribution shifts.

\subsection{Mathematical Formulation}
We now formalize the Routing-driven Transformer (RwF) layer. Let $Z_\ell \in \mathbb{R}^{L \times d}$ denote the backbone token embeddings \emph{input to layer $\ell$}. Before applying the self-attention block, we compute input-conditioned routing prompts through an associative operator $\mathcal{H}$ that generates $P_\ell \in \mathbb{R}^{m \times d}$, where $m \ll L$.

Let $Q_\ell \in \mathbb{R}^{m \times d}$ be learnable query vectors. Define projections
\begin{equation}
K_\ell = Z_\ell W_K, 
\quad 
V_\ell = Z_\ell W_V, 
\quad 
\tilde Q_\ell = Q_\ell W_Q.
\label{eq:learnable_params}
\end{equation}

The associative routing operator is defined as
\begin{equation}
\label{eq:associative_op}
P_\ell 
= 
\mathcal{H}(Q_\ell, Z_\ell) 
= 
\rho
\!\left(
\beta \tilde Q_\ell K_\ell^\top
\right) V_\ell,
\end{equation}

where $\rho$ denotes the row-wise softmax operator and $\beta > 0$ is the inverse temperature.

The routing matrix
\[
A_\ell(Z_\ell) = 
\rho
\
\!\left(
\beta \tilde Q_\ell K_\ell^\top
\right)
\in \mathbb{R}^{m \times L}
\]
defines, for each query, a probability distribution over the $L$ input tokens.
The retrieved prompts are convex combinations of the input features:
\[
P_\ell = A_\ell(Z_\ell) V_\ell.
\]

The retrieved prompts are then concatenated with the original input sequence to form
\[
R_\ell = [P_\ell; Z_\ell],
\]
which is processed by the self-attention block of layer $\ell$ that updates $R_\ell$ to $\tilde R_\ell = [\tilde P_\ell; \tilde Z_\ell]$. From the attended tokens $\tilde R_\ell$, only the backbone token representations $\tilde Z_\ell$ are propagated to the next MLP blocks, while the transformed routing prompts $\tilde P_\ell$ are discarded. Discarding $\tilde{P}_\ell$ prevents accumulation of task-specific prompt states; their influence is fully mediated through the attention update of $Z_\ell$, ensuring that routing remains input-driven rather than stored across tasks.

Importantly, Eq.~\ref{eq:associative_op} is computed in closed form during the forward pass. While parameters are optimized through gradient descent across the stream, routing decisions themselves do not require iterative search or repeated exposure to the same data.

\paragraph{Energy-Based Interpretation.}
Following the modern Hopfield formulation~\cite{ramsauer2021hopfield}, Eq.~\ref{eq:associative_op} admits a variational interpretation. For a query $q \in Q_\ell$, define $\tilde q = q W_Q$ and let $k_i$ denote rows of $K_\ell$. Retrieval corresponds to minimizing
\[
\mathcal F(p; q)
=
- \sum_{i=1}^L p_i \langle \tilde q, k_i \rangle
+ \beta^{-1} H(p),
\]
where $H(p)$ is Shannon entropy.

The alignment term encourages the routing distribution to concentrate on tokens most compatible with the current feature geometry (\emph{plasticity}), while the entropy term discourages degenerate one-hot assignments and promotes smoother allocations (\emph{stability}). Because negative entropy is strictly convex over the probability simplex, the energy $\mathcal F$ admits a unique global minimizer $p^*$, given in closed form by the softmax distribution in Eq.~\ref{eq:associative_op}. Under this view, routing in RwF corresponds to the equilibrium of a strictly convex energy functional for fixed parameters, obtained directly in a single forward pass without iterative optimization. In OCL, this enables routing to adapt immediately to distribution shifts, reconfiguring internal representations before gradient-based parameter updates accumulate.

The energy-based formulation is mathematically equivalent to the softmax retrieval used in attention; its role here is simply to interpret routing as a convex, smooth, input-conditioned mapping governing how representations are dynamically redistributed within the transformer.

\begin{figure*}[t]
    \centering
    \includegraphics[width=\linewidth]{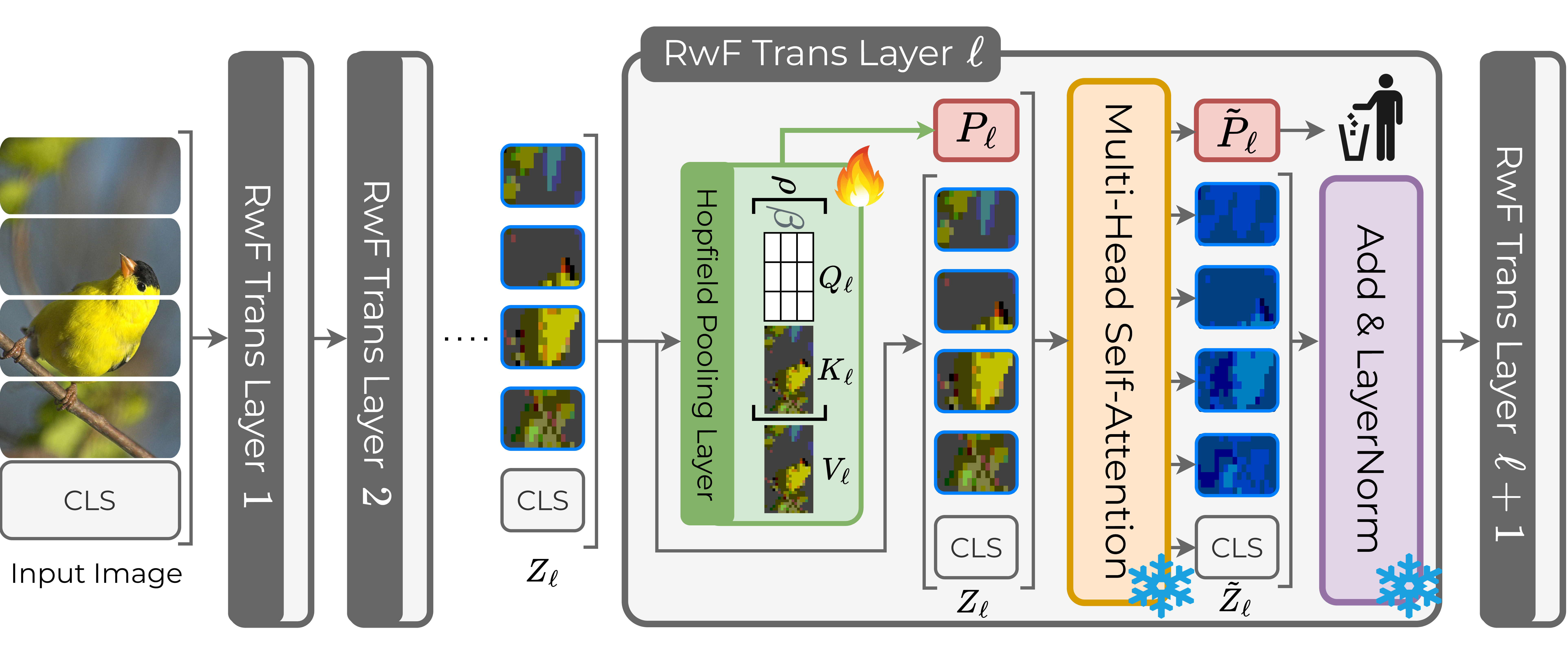}
    \caption{\textbf{RwF layer: routing-augmented transformer block.} Given input tokens $Z_\ell$, a Hopfield-based associative retrieval module generates input-conditioned routing prompts $P_\ell$ via energy-based pooling over token features. The retrieved $P_\ell$ are concatenated with the $Z_\ell$ tokens and processed by the standard Multi-Head Self-Attention (MHSA). After MHSA, only the backbone tokens  $\tilde{Z}_\ell$ are propagated to the MLP blocks and then to the next RwF transformer layer $\ell+1$, while $\tilde{P}_\ell$ are discarded.}
    \label{fig:method}
\end{figure*}

\paragraph{Architectural Smoothness.}
The operator $\mathcal{H}$ is continuous in $Z_\ell$, being composed of linear projections and softmax normalization. LayerNorm in transformers keeps token norms bounded and ensuring that pre-softmax similarity scores lie in a compact domain. Under this condition, softmax is Lipschitz, giving
\[
\|A_\ell(Z_\ell + \Delta Z_\ell) - A_\ell(Z_\ell)\|
\le
L_\beta \|\Delta Z_\ell\|,
\]
for some constant $L_\beta$.

Thus, small feature shifts induce proportionally small changes in routing weights, providing a structural buffer against the abrupt representation jumps that often cause catastrophic forgetting. This smoothness is useful in OCL: as parameters evolve across tasks, routing changes only gradually, avoiding sudden reallocation of attention patterns. In prompt-based methods, misaligned routing typically requires several gradient steps to correct; in contrast, RwF recomputes routing from the current input at every forward pass. Routing therefore adapts immediately to distribution shifts, decoupling its timescale from that of parameter optimization.

\subsection{Online Continual Learning via Hopfield Pooling}

For online continual learning, we instantiate the associative operator $\mathcal{H}$ using a HopfieldPooling layer, which is algebraically equivalent to Eq.~\ref{eq:associative_op}. Unlike standard self-attention ($L \rightarrow L$), HopfieldPooling performs a many-to-few mapping ($L \rightarrow m$), compressing the current token sequence into a small set of input-conditioned routing prompts.  This design is crucial: routing is recomputed analytically at every forward pass, allowing the model to reconfigure internal representations immediately upon observing new data. 
Thus, the RwF transformer layer is defined by instantiating $\mathcal{H}$ with HopfieldPooling and is illustrated in Fig.~\ref{fig:method}. No task-specific modules are allocated, and no replay buffer or iterative prompt specialization is required. Instead, adaptation occurs through closed-form associative retrieval directly within the backbone.
The projection matrices used for routing are decoupled from the backbone attention layers, ensuring that routing dynamics can adapt without disrupting pre-trained feature extraction. As we show in the ablation study (Sec.~\ref{sec:results}), inserting a small number of HopfieldPooling layers in the early transformer blocks yields strong performance while introducing only a minimal fraction of additional parameters. $W_K$ and $W_V$ are untrained; indeed, updating these projections causes the similarity space used by $\mathcal{H}$ to drift as the backbone changes, leading to unstable routing; keeping them fixed provides a stationary basis so that variations in $A_\ell(Z)$ primarily reflect the input. Moreover, experimental results confirmed that these projections do not yield measurable gains, indicating that a fixed projection basis is sufficient for stable routing in OCL. 

Overall, the novelty of RwF lies in introducing a many-to-few associative routing mechanism within each transformer block. Unlike cross-attention with fixed learnable queries, RwF recomputes $m$ input-conditioned routing prompts analytically at every forward pass from the current token geometry. In contrast to prior Hopfield-based memory modules, retrieval operates inside the backbone to dynamically reallocate representational subspaces rather than storing external patterns or task-specific prompts. This enables routing to adapt immediately to distribution shifts without expanding the backbone or introducing task-dependent parameters.

\section{Performance Analysis}
\subsection{Benchmarks}

\noindent \textbf{Datasets.} We evaluate RwF on three Class-IL benchmarks following~\cite{onlinelora}: \emph{Split-CIFAR100}~\cite{krizhevsky2009learning}, \emph{Split-ImageNet-R}~\cite{hendrycks2021many}, and \emph{Split-ImageNet-S}~\cite{gao2022luss}. Data are presented as a stream of tasks with disjoint class subsets and, at test time, the model predicts over the union of all classes observed so far. For \emph{Split-CIFAR100}, we adopt a 10-task protocol with 10 classes per task. For \emph{Split-ImageNet-R}, we use 10 tasks with 20 classes per task. For \emph{Split-ImageNet-S}, we consider the full 1000-class label space split into 10 tasks, with 100 classes per task. 

\noindent \textbf{Metrics.} We report two standard CL metrics. The first is the \textit{Final Average Accuracy} (\(\mathrm{A}_{\text{Final}} (\uparrow) \)) as $\frac{1}{T}\sum_{i=1}^{T}{a_i^T}$, 
where $a_i^T$ is the accuracy of the final model computed after learning the last task.

The second is \textit{Forgetting} (\(\downarrow\)), which measures how much performance on past tasks deteriorates as new tasks are learned. We define it as the average gap between the best accuracy achieved on each task during training and its final accuracy after learning all tasks.

\subsection{Training procedure and evaluation protocol}

We compare RwF against state-of-the-art approaches designed/adapted for strict OCL, including parameter-efficient, replay-based, and prompt-based methods.

All experiments follow the single-pass evaluation protocol introduced in~\cite{onlinelora}, using a ViT-B/16 backbone pre-trained on ImageNet-21k, Adam optimizer, and batch size 64. Consistent with the OCL setting, each task is processed for a single epoch (i.e., each sample is observed exactly once). For rehearsal-based methods, the memory buffer is fixed to 500 exemplars across all experiments\footnote{Buffer-based results follow the configuration of~\cite{onlinelora}, which we replicate exactly.}. We use $m = 30$ for the dimension of query $\tilde{Q}_\ell \in \mathbb{R}^{m \times d}$ and set $\beta=1/\sqrt{d}$.

For fairness and reproducibility, we restrict comparisons to methods with publicly available implementations. All baselines were executed using official codebases under a unified experimental protocol. Importantly, we performed dedicated hyperparameter tuning for each method under the strict online regime. Hyperparameters reported in the original papers—often optimized for multi-epoch or partially offline protocols—led to substantially degraded performance when directly applied to single-pass OCL. We therefore conducted controlled grid searches to ensure competitive configurations. Detailed comparisons between original and tuned settings are provided in the supplementary material. All results are reported as mean $\pm$ standard deviation over three runs.

\subsection{Results}\label{sec:results}

We design the empirical evaluation to validate the core hypotheses introduced in Sec.~\ref{sec:method}. 
First, we assess whether energy-based associative routing improves OCL performance (Table~\ref{tab:vitb16_results}). Since RwF is a \emph{buffer-free} method, our primary comparisons focus on prompt- and adapter-based approaches, while replay-based methods are also reported to contextualize results relative to broader state-of-the-art baselines. 
\begin{table*}[t]
\centering
\small
\setlength{\tabcolsep}{3pt}
\caption{\textbf{OCL performance comparison.} Final Average Accuracy $A_{\text{Final}}(\uparrow)$ and Forgetting $(\downarrow)$ on Split-CIFAR100, Split-ImageNet-R, and Split-ImageNet-S. Best: bold, second: underline, third: italic. RwF achieves top performance on large-scale ImageNet benchmarks while ranking competitively on CIFAR100.}

\resizebox{\textwidth}{!}{%
\begin{tabular}{lcccccc}
\toprule
 & \multicolumn{2}{c}{\textbf{Split-CIFAR100}} &
   \multicolumn{2}{c}{\textbf{Split-ImageNet-R}} &
   \multicolumn{2}{c}{\textbf{Split-ImageNet-S}} \\
\cmidrule(r){2-3}\cmidrule(r){4-5}\cmidrule(r){6-7}
\multirow{1}{*}{Method} &
$A_{\text{Final}}\,(\uparrow)$ & Forgetting $(\downarrow)$ &
$A_{\text{Final}}\,(\uparrow)$ & Forgetting $(\downarrow)$ &
$A_{\text{Final}}\,(\uparrow)$ & Forgetting $(\downarrow)$ \\
\midrule
\textit{Joint}         & 89.50{\scriptsize$\pm$0.04} & \multicolumn{1}{c}{--} &
                      76.78{\scriptsize$\pm$0.44} & \multicolumn{1}{c}{--} &
                      63.82{\scriptsize$\pm$0.02} & \multicolumn{1}{c}{--} \\
\rowcolor{gray!10}
\textit{Finetune}       & 15.83{\scriptsize$\pm$0.25} & 79.68{\scriptsize$\pm$0.08} &
                      12.06{\scriptsize$\pm$0.34} & 79.61{\scriptsize$\pm$0.04} &
                      7.09{\scriptsize$\pm$0.08} & 26.45{\scriptsize$\pm$0.14} \\

\midrule
AGEM~\cite{agem}    & 12.67{\scriptsize$\pm$1.87} & 82.51{\scriptsize$\pm$2.27} &
                       5.60{\scriptsize$\pm$2.42} & 53.97{\scriptsize$\pm$1.97} &
                       0.16{\scriptsize$\pm$0.04} &  9.42{\scriptsize$\pm$0.17} \\
\rowcolor{gray!10}
ER~\cite{er}       & 44.85{\scriptsize$\pm$1.83} & 44.67{\scriptsize$\pm$4.29} &
                      40.99{\scriptsize$\pm$3.96} & 32.38{\scriptsize$\pm$0.89} &
                      30.21{\scriptsize$\pm$0.70} & 37.14{\scriptsize$\pm$1.83} \\
EWC++~\cite{ewcpp}  & 10.61{\scriptsize$\pm$0.74} & 84.10{\scriptsize$\pm$1.11} &
                       3.86{\scriptsize$\pm$2.02} & 36.95{\scriptsize$\pm$1.46} &
                       0.32{\scriptsize$\pm$0.28} & 22.46{\scriptsize$\pm$4.69} \\
\rowcolor{gray!10}
MIR~\cite{mir}      & 48.36{\scriptsize$\pm$1.33} & 43.41{\scriptsize$\pm$1.02} &
                      41.51{\scriptsize$\pm$2.99} & 31.32{\scriptsize$\pm$5.17} &
                      30.33{\scriptsize$\pm$3.21} & 33.82{\scriptsize$\pm$1.03} \\
GDumb~\cite{gdumb}  & 41.00{\scriptsize$\pm$19.97} & \multicolumn{1}{c}{--} &
                       8.87{\scriptsize$\pm$3.16}  & \multicolumn{1}{c}{--} &
                       1.65{\scriptsize$\pm$0.22}  & \multicolumn{1}{c}{--} \\
\rowcolor{gray!10}
PCR~\cite{pcr}      & 48.40{\scriptsize$\pm$0.15} & 46.23{\scriptsize$\pm$1.29} &
                      46.11{\scriptsize$\pm$3.03} & 25.50{\scriptsize$\pm$4.01} &
                      38.75{\scriptsize$\pm$0.22} & 35.01{\scriptsize$\pm$1.12} \\
DER++~\cite{der} & 49.08{\scriptsize$\pm$1.15} & 56.94{\scriptsize$\pm$4.72} &
                      39.67{\scriptsize$\pm$4.23} & 24.26{\scriptsize$\pm$3.64} &
                       6.47{\scriptsize$\pm$0.46} & 15.34{\scriptsize$\pm$0.15} \\
\rowcolor{gray!10}
LODE(DER++)~\cite{lode} &
                      44.29{\scriptsize$\pm$1.48} & 45.54{\scriptsize$\pm$3.32} &
                      36.60{\scriptsize$\pm$2.46} & 20.40{\scriptsize$\pm$2.06} &
                       9.97{\scriptsize$\pm$2.29} &  8.48{\scriptsize$\pm$1.24} \\
EMA(DER++)~\cite{ema} &
                      42.28{\scriptsize$\pm$3.46} & 55.59{\scriptsize$\pm$1.48} &
                      41.75{\scriptsize$\pm$1.98} & 32.65{\scriptsize$\pm$1.55} &
                      16.88{\scriptsize$\pm$2.23} & 36.28{\scriptsize$\pm$1.09} \\
\rowcolor{gray!10}
EMA(RAR)~\cite{ema} &
                      47.10{\scriptsize$\pm$0.82} & 50.01{\scriptsize$\pm$0.35} &
                      48.10{\scriptsize$\pm$0.33} & 39.36{\scriptsize$\pm$0.04} &
                      14.06{\scriptsize$\pm$0.37} & 28.09{\scriptsize$\pm$3.25} \\
Online-Lora~\cite{onlinelora}       & 49.40{\scriptsize$\pm$1.36} & 41.74{\scriptsize$\pm$2.35} &
                      48.18{\scriptsize$\pm$6.03} & 23.85{\scriptsize$\pm$1.48} &
                      47.06{\scriptsize$\pm$0.24} & 26.09{\scriptsize$\pm$0.34} \\
\rowcolor{gray!10}
L2P~\cite{l2p}                 & 78.11{\scriptsize$\pm$0.25} & 7.98{\scriptsize$\pm$0.25} &
                      57.97{\scriptsize$\pm$0.31} & 6.76{\scriptsize$\pm$0.23} &
                      31.45{\scriptsize$\pm$0.58} & 9.43{\scriptsize$\pm$0.53} \\
CODA-Prompt~\cite{coda}          & 77.28{\scriptsize$\pm$0.56} & 11.87{\scriptsize$\pm$0.64} &
                      \underline{66.16{\scriptsize$\pm$0.39}} & 8.95{\scriptsize$\pm$0.34} &
                      47.59{\scriptsize$\pm$0.53} & 5.73{\scriptsize$\pm$0.67} \\
\rowcolor{gray!10}
DualPrompt~\cite{dualprompt}          & \underline{83.58{\scriptsize$\pm$0.28}} & 6.33{\scriptsize$\pm$0.71} &
                      60.88{\scriptsize$\pm$0.62} & 6.15{\scriptsize$\pm$0.33} &
                      42.40{\scriptsize$\pm$0.34} &               8.35{\scriptsize$\pm$0.89} \\
InfLoRA~\cite{inflora}          & 81.08{\scriptsize$\pm$1.77} & 6.90{\scriptsize$\pm$1.84} &
                      \textit{62.20{\scriptsize$\pm$0.87}} & 6.60{\scriptsize$\pm$0.38} &
                      53.83{\scriptsize$\pm$0.08} & 3.89{\scriptsize$\pm$0.23} \\
\rowcolor{gray!10}
APER + Adapt.~\cite{aper}          & 81.35{\scriptsize$\pm$0.09} & 4.87{\scriptsize$\pm$0.27} &
                      55.12{\scriptsize$\pm$0.02} & 6.48{\scriptsize$\pm$1.55} &
                      \underline{56.26{\scriptsize$\pm$0.37}} & 5.11{\scriptsize$\pm$0.24} \\
EASE~\cite{ease}          & \textbf{84.81{\scriptsize$\pm$1.19}} & 4.86{\scriptsize$\pm$0.38} &
                      55.44{\scriptsize$\pm$0.24} & 7.20{\scriptsize$\pm$0.04} &
                      \textit{55.89{\scriptsize$\pm$0.19}} & 3.96{\scriptsize$\pm$0.44} \\
\midrule
\textbf{RwF}            & \textit{82.48{\scriptsize$\pm$0.31}} & 6.71{\scriptsize$\pm$0.43}&
                      \textbf{74.09{\scriptsize$\pm$0.29}} & 5.98{\scriptsize$\pm$0.73} &
                      \textbf{61.37{\scriptsize$\pm$0.43}} & 7.23{\scriptsize$\pm$0.35} \\
\addlinespace[0.2em]

\bottomrule
\end{tabular}%
}
\label{tab:vitb16_results}
\end{table*}

Second, we test robustness under limited optimization opportunities by evaluating few-shot and data-scarce regimes (Table~\ref{tab:fewshot_ablation_split_imagenet_r}). Third, we analyze scalability under increasing task fragmentation as the number of sequential tasks grows (Fig.~\ref{fig:tasks_ablation_split_imagenet_r}). Finally, we examine the architectural role of routing through depth ablations (Table~\ref{tab:ablation_hp_imagenetr}), assessing how early-layer associative retrieval affects stability–plasticity dynamics.\\ \vspace{-0.7cm}

\subsubsection{OCL Performance.}
\label{sec:cil-performance}
We first evaluate whether replacing parameter-specialization strategies with associative routing improves performance under the OCL regime. Table~\ref{tab:vitb16_results} compares RwF against state-of-the-art methods, including replay-based (ER~\cite{er}, MIR~\cite{mir}, DER++~\cite{der}), regularization-based (EWC++~\cite{ewcpp}), prompt-based (L2P~\cite{l2p}, DualPrompt~\cite{dualprompt}, CODA-Prompt~\cite{coda}), and adapter-based approaches (Online-LoRA~\cite{onlinelora}, InfLoRA~\cite{inflora}, APER~\cite{aper}, EASE~\cite{ease}). We also report the \emph{upper bound} (Joint training on all data) and the \emph{lower bound} (sequential fine-tuning without mitigation) for reference.
Notably, among adapter-based methods, Online-LoRA~\cite{onlinelora} relies on a small buffer of hard samples for parameter-importance estimation, InfLoRA~\cite{inflora} maintains multiple task-specific LoRA modules, EASE~\cite{ease} requires task-orthogonal initialization, and APER~\cite{aper} employs dual ViT backbones (pre-trained and fine-tuned replicas) to decouple plastic and stable representations. Similarly, prompt-based approaches such as L2P~\cite{l2p}, DualPrompt~\cite{dualprompt}, and CODA-Prompt~\cite{coda} allocate or select prompts conditioned on explicit task boundaries during training, relying on task identity to regulate specialization.

In contrast, RwF operates within a single backbone, requires no replay buffer, no parameter replication, no special initialization, and no task-boundary information. Adaptation emerges through input-conditioned associative routing computed at every forward pass, making the method both competitive in accuracy and better aligned with realistic OCL settings where task transitions are not explicitly signaled.

On Split-ImageNet-R and Split-ImageNet-S, RwF achieves substantial improvements, reaching \textbf{74.09\%} and \textbf{61.37\%}, respectively. It outperforms DualPrompt (60.88\% / 42.40\%), CODA-Prompt (66.16\% / 47.59\%), InfLoRA (62.20\% / 53.83\%), and EASE (55.44\% / 55.89\%) by clear margins. RwF also surpasses APER (55.12\% / 56.26\%), despite APER employing a duplicated backbone to decouple plastic and stable representations. These results suggest that input-conditioned associative routing within a single backbone can provide stronger interference mitigation than task-related parameters or dual-backbone strategies when the class space is large and semantically diverse.

On Split-CIFAR100, EASE attains the highest accuracy (84.81\%), followed by DualPrompt (83.58\%) and RwF (82.48\%). Although RwF does not achieve the top score in this setting, it remains competitive with other parameter-efficient methods. The smaller gap compared to the ImageNet benchmarks is consistent with CIFAR100’s low resolution and limited spatial detail. Since associative routing operates by reweighting token-level representations based on feature similarity, its effectiveness depends on the richness of intermediate feature geometry. In low-resolution images, where fine-grained spatial structure is compressed and semantic variability across tasks is more limited, the representational space offers fewer informative routing directions, reducing the relative advantage of dynamic feature reallocation.

In the next two analyses we focus on Split-ImageNet-R. Results on Split-CIFAR100 and Split-ImageNet-S are included in the supplementary material.

\subsubsection{Robustness in Few-Shot Regimes.}
We next evaluate performance under progressively reduced training data, simulating increasingly constrained online conditions. Table~\ref{tab:fewshot_ablation_split_imagenet_r} reports Final Average Accuracy on Split-ImageNet-R as the available training samples per task decrease from 100\% to 20\%.

\begin{table*}[htb]
\centering
\setlength{\tabcolsep}{4pt}
\caption{\textbf{Performance under few-shot data regimes (Split-ImageNet-R).}
Final Average Accuracy ($A_{\text{Final}}(\uparrow)$) as the fraction of available training samples decreases from 100\% to 20\%. Best: bold, second: underline, third: italic. RwF maintains the highest accuracy across all evaluated data regimes.}
\label{tab:fewshot_ablation_split_imagenet_r}
\resizebox{0.7\textwidth}{!}{%
\begin{tabular}{lcccc}
\toprule
 & \multicolumn{4}{c}{\textbf{Split-ImageNet-R}} \\
\cmidrule(r){2-5}
\textbf{Method} & 100\% & 70\% & 50\% & 20\% \\
\midrule
L2P~\cite{l2p}                & 57.97{\scriptsize$\pm$0.31} & 56.15{\scriptsize$\pm$0.13} & 53.71{\scriptsize$\pm$0.32} & 44.79{\scriptsize$\pm$1.14} \\
\rowcolor{gray!10}
CODA-Prompt~\cite{coda}         & \underline{66.16{\scriptsize$\pm$0.39}} & \underline{64.60{\scriptsize$\pm$1.04}} & \underline{61.63{\scriptsize$\pm$0.28}} & \underline{54.79{\scriptsize$\pm$1.16}} \\
DualPrompt~\cite{dualprompt}         & 60.88{\scriptsize$\pm$0.62} & \textit{58.67{\scriptsize$\pm$0.81}} & \textit{56.87{\scriptsize$\pm$0.33}} & 47.68{\scriptsize$\pm$0.52} \\
\rowcolor{gray!10}
InfLoRA~\cite{inflora}            & \textit{62.20{\scriptsize$\pm$0.87}} & 51.94{\scriptsize$\pm$0.67} & 37.90{\scriptsize$\pm$0.98} & 6.65{\scriptsize$\pm$0.08} \\
APER~\cite{aper}               & 55.12{\scriptsize$\pm$0.02} & 55.04{\scriptsize$\pm$0.27} & 54.16{\scriptsize$\pm$0.45} & 49.72{\scriptsize$\pm$0.13} \\
\rowcolor{gray!10}
EASE~\cite{ease}               & 55.44{\scriptsize$\pm$0.24} & 54.36{\scriptsize$\pm$0.47} & 53.89{\scriptsize$\pm$0.49} & \textit{49.76{\scriptsize$\pm$0.07}} \\
\midrule
\textbf{RwF}       & \textbf{74.09{\scriptsize$\pm$0.29}} & \textbf{71.89{\scriptsize$\pm$0.52}} & \textbf{69.54{\scriptsize$\pm$0.87}} & \textbf{62.29{\scriptsize$\pm$0.75}} \\
\bottomrule
\end{tabular}%
}
\end{table*}

RwF consistently achieves the highest accuracy across all data regimes. While all methods degrade as supervision becomes scarce, the performance drop of RwF is substantially more controlled. For instance, when reducing the data to 20\%, prompt-based methods such as CODA-Prompt and DualPrompt exhibit significant declines, and InfLoRA collapses sharply. In contrast, RwF maintains 62.29\%, preserving a clear margin over all competitors.

This behavior is coherent with the architectural design of RwF. Prompt- and adapter-based approaches rely on iterative gradient-driven specialization of additional parameters; when the number of samples per task is reduced, these parameters receive insufficient updates to form stable task-specific representations. Associative routing, instead, performs input-conditioned reallocation of feature subspaces within each forward pass. Because routing decisions are computed analytically from the current feature geometry, adaptation does not rely exclusively on repeated exposure to task data.

\subsubsection{Incremental Learning Scalability.}
We next analyze how performance evolves as the number of sequential tasks $t$ increases. When the total label space is fixed, increasing the number of tasks reduces the number of classes per task and induces more frequent distribution shifts. This setting amplifies representational interference: the model must repeatedly reorganize feature subspaces under progressively finer task fragmentation. Methods that rely on task-specific parameter specialization may struggle in this regime, as accumulated modular separation becomes harder to maintain when the number of competing tasks grows.

\begin{figure}[h]
  \centering
  \includegraphics[width=0.7\textwidth]{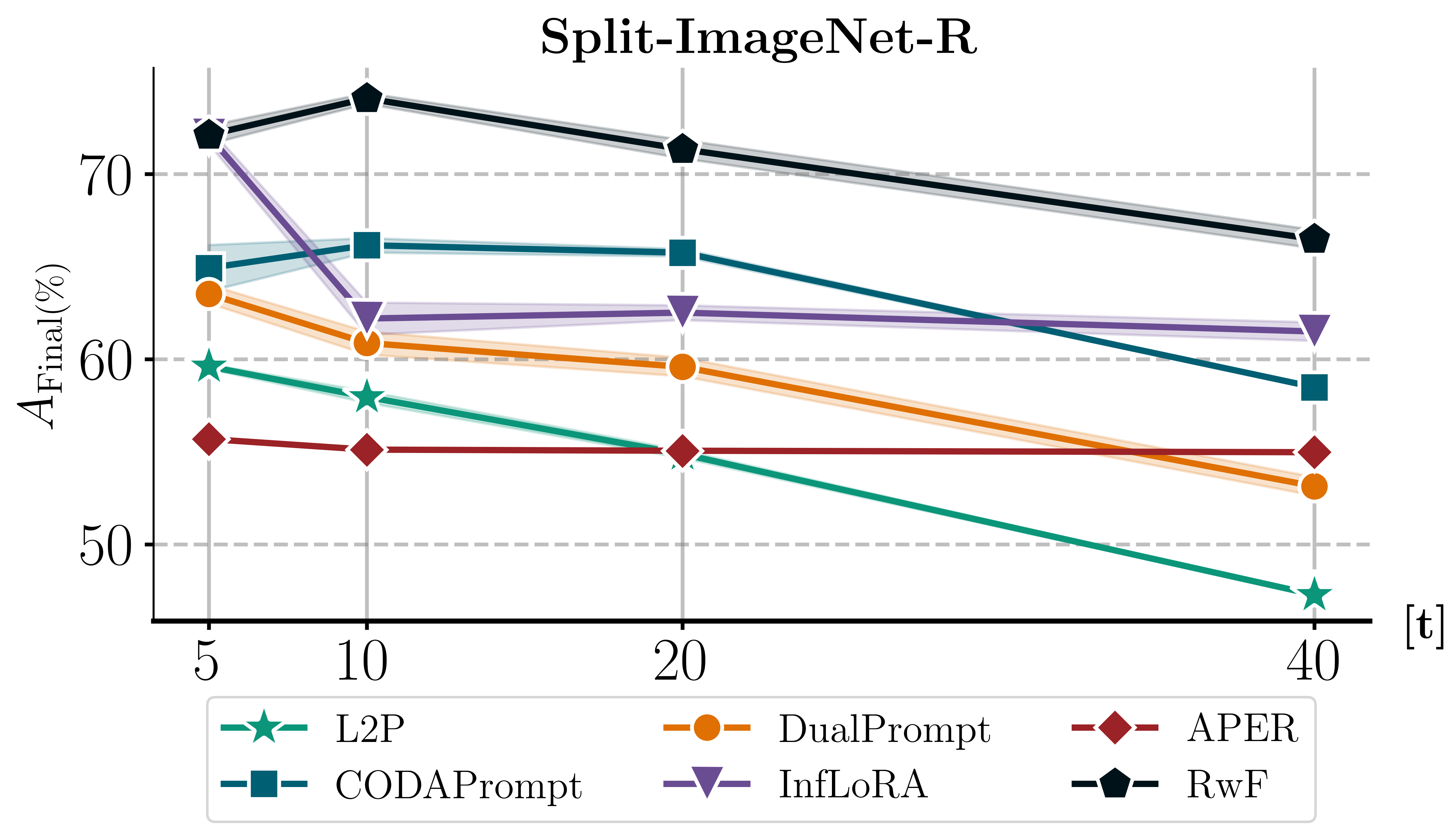}
  \caption{\textbf{Performance scaling with increasing task fragmentation (Split-ImageNet-R).}
  Final Average Accuracy ($\mathrm{A}_{\text{Final}}$) $(\uparrow)$ as the number of sequential tasks $t$ increases from 5 to 40.}
  \label{fig:tasks_ablation_split_imagenet_r}
\end{figure}

Fig.~\ref{fig:tasks_ablation_split_imagenet_r} reports Final Average Accuracy on Split-ImageNet-R as the number of tasks increases from 5 to 40. As expected, all methods exhibit performance degradation with finer task splits. However, RwF consistently outperforms prompt- and adapter-based baselines across all task counts. While L2P and DualPrompt show steady declines under increasing fragmentation, RwF preserves a consistent margin, achieving 66.49\% even with 40 tasks.

This behavior is coherent with the routing formulation: RwF recomputes routing distributions from the current feature geometry at every forward pass. Representation allocation is therefore continuously re-evaluated rather than tied to historical gradient-driven specialization. As task boundaries become more frequent, this dynamic reconfiguration becomes increasingly advantageous. Notably, APER exhibits relative stability as task count increases. This likely stems from its similarity-based matching against compact class prototypes, which reduces inter-prototype interference when fewer classes are present per task. Nevertheless, its absolute performance remains consistently below that of RwF.\\ \vspace{-0.7cm}

\subsubsection{Effect of Routing Depth and Capacity.}
We investigate how the number and position of HopfieldPooling (HP) layers influence performance. If associative routing modifies internal representation dynamics, its effect should depend on how many routing layers are introduced and where they operate within the transformer. Table~\ref{tab:ablation_hp_imagenetr} reports results on Split-CIFAR100, Split-ImageNet-R, and Split-ImageNet-S as we vary the number of HP layers $k$, along with the percentage of learnable parameters relative to the backbone (excluding the classification head). Routing layers are inserted either in the first $k$ blocks (First-$k$) or in the last $k$ blocks (Last-$k$).
\begin{table*}[htb!]
\centering
\footnotesize
\setlength{\tabcolsep}{3.5pt}
\caption{\textbf{Effect of routing depth on Split-CIFAR100, Split-ImageNet-R and Split-ImageNet-S.}
Final Average Accuracy ($\mathrm{A}_{\text{Final}} (\uparrow)$) as the number of HopfieldPooling (HP) layers $k$ increases.
\emph{First-$k$} inserts HP layers in the first $k$ transformer block; \emph{Last-$k$} inserts them in last $k$ blocks ($12\dots12-k+1$). Best: bold, second: underline, third: italic. \colorbox{green!20}{Green cells} indicate chosen configuration used for all paper results.}
\resizebox{0.8\textwidth}{!}{%
\begin{tabular}{c c cc cc cc}
\toprule
\multirow{2}{*}{$k$} & \multirow{2}{*}{\%} &
\multicolumn{2}{c}{Split-CIFAR100} &
\multicolumn{2}{c}{Split-ImageNet-R} &
\multicolumn{2}{c}{Split-ImageNet-S} \\
\cmidrule(lr){3-4}\cmidrule(lr){5-6}\cmidrule(lr){7-8}
& & First-$k$ & Last-$k$ & First-$k$ & Last-$k$ & First-$k$ & Last-$k$ \\
\midrule
0 & 0.0 & \multicolumn{2}{c}{15.83{\scriptsize$\pm$0.25}} &
\multicolumn{2}{c}{12.06{\scriptsize$\pm$0.34}} &
\multicolumn{2}{c}{7.09{\scriptsize$\pm$0.08}} \\
\rowcolor{gray!10}
1  & 0.7 & 79.57{\scriptsize$\pm$0.54} & 65.98{\scriptsize$\pm$0.73} & 67.78{\scriptsize$\pm$1.02} & 60.10{\scriptsize$\pm$0.61} & 58.36{\scriptsize$\pm$0.55} &  49.19{\scriptsize$\pm$0.60}\\
\cellcolor{green!20}3  & \cellcolor{green!20}2.1 & \cellcolor{green!20} \underline{82.48{\scriptsize$\pm$0.31}} & 60.81{\scriptsize$\pm$4.12} & \cellcolor{green!20}\textit{74.09{\scriptsize$\pm$0.29}} & 55.92{\scriptsize$\pm$0.61} & \cellcolor{green!20}61.37{\scriptsize$\pm$0.43} & 48.88{\scriptsize$\pm$0.62} \\
\rowcolor{gray!10}
5  & 3.5 & \textbf{82.89{\scriptsize$\pm$0.47}} & 61.56{\scriptsize$\pm$3.16} & \underline{74.68{\scriptsize$\pm$0.04}} & 60.64{\scriptsize$\pm$2.00} & 62.19{\scriptsize$\pm$0.48} &  51.83{\scriptsize$\pm$0.29} \\
7  & 4.9 & \textit{81.65}{\scriptsize$\pm$1.06} & 64.77{\scriptsize$\pm$0.59} & \textbf{75.39{\scriptsize$\pm$0.17}} & 63.95{\scriptsize$\pm$0.52} & \textit{62.88{\scriptsize$\pm$0.57}} &  53.83{\scriptsize$\pm$0.42} \\
\rowcolor{gray!10}
10 & 7.1 & 74.76{\scriptsize$\pm$0.82} & 70.40{\scriptsize$\pm$0.76} & 73.68{\scriptsize$\pm$0.80} & 65.01{\scriptsize$\pm$1.12} & \underline{63.10{\scriptsize$\pm$0.52}} & 58.35{\scriptsize$\pm$0.72} \\
12 & 8.5 &
\multicolumn{2}{c}{69.64{\scriptsize$\pm$2.81}} &
\multicolumn{2}{c}{65.29{\scriptsize$\pm$0.42}} &
\multicolumn{2}{c}{\textbf{63.68{\scriptsize$\pm$0.42}}} \\
\bottomrule
\end{tabular}%
}
\label{tab:ablation_hp_imagenetr}
\end{table*}

Across datasets, inserting HP layers in early blocks (First-$k$) generally yields stronger performance than restricting routing to deeper layers, especially for small and moderate $k$. On Split-CIFAR100, accuracy improves up to $k=5$ (82.89\%) before declining. On Split-ImageNet-R, performance increases up to $k=7$ (75.39\%) and then saturates. On Split-ImageNet-S, the best result is achieved with full-depth routing ($k=12$), albeit with a substantially larger parameter budget. These trends suggest that interference is most effectively mitigated at shared, lower-level representations. Early-layer routing allows conflicts to be intercepted before propagating to deeper, more task-specific features. Balancing accuracy and parameter efficiency, we adopt $k=3$ HP layers in the first blocks as our default configuration. This setting provides strong performance with only $\sim$2.1\% additional trainable parameters. Larger $k$ yields limited gains at significantly higher cost, while smaller $k$ underutilizes routing.

\subsubsection{Computational Cost and Parameter Efficiency.} Beyond accuracy, it is important to assess whether the gains of RwF stem from increased model capacity or computational overhead. 

\begin{table}[h]
  \centering
  \renewcommand{\arraystretch}{1.3}
  \caption{\textbf{Trainable parameters (\%) relative to the full ViT backbone.} RwF maintains a low parameter footprint, comparable to other parameter-efficient CL methods.}
  \begin{tabular}{l c}
    \toprule
    \textbf{Model} & \textbf{\# params}  \\
    \midrule
    L2P~\cite{l2p}              & 0.47\% \\
    \rowcolor{gray!10}
    DualPrompt~\cite{dualprompt}& 0.58\% \\
    CODA-Prompt~\cite{coda}      & 5.00\% \\
    \rowcolor{gray!10}
    InfLoRA~\cite{inflora}      & 1.70\% \\
    APER~\cite{aper}            & 1.50\% \\
    \rowcolor{gray!10}
    EASE~\cite{ease}            & 1.32\% \\
    \midrule
    \textbf{RwF}                & 2.13\% \\
    \bottomrule
  \end{tabular}
  \label{tab:param_percent}
\end{table}

Table~\ref{tab:param_percent} shows that RwF introduces only 2.13\% additional trainable parameters, comparable to or lower than competitors, and significantly below compositional prompt methods such as CODA-Prompt. The improvements observed on Split-ImageNet-R and Split-ImageNet-S therefore cannot be attributed to parameter scaling alone. For the query $\tilde{Q}_\ell \in \mathbb{R}^{m \times d}$, we tested different values of $m$, as it directly affects the routing capacity of a HopfieldPooling layer. Increasing this parameter consequently increases the number of trainable parameters. We decided to use $m = 30$, which yielded satisfactory accuracy scores while keeping the number of additional parameters manageable. From a computational perspective, associative routing introduces a many-to-few attention-like mapping of complexity $\mathcal{O}(mL d)$ per layer, where $L$ is the token length, $d$ the feature dimension, and $m \ll L$ the number of routing prompts. Since $m$ is small and routing is applied only in a subset of transformer blocks, the additional cost remains modest compared to self-attention ($\mathcal{O}(L^2 d)$). 

\section{Limitations}

RwF performs less effectively on fine-grained classification benchmarks (see supplementary material), such as CUB-200~\cite{wah2011caltech}, where class distinctions rely on subtle, localized cues rather than coarse distributional shifts. The HopfieldPooling routing mechanism aggregates token features through similarity-weighted combinations, which may smooth highly discriminative local details. Moreover, because fine-grained classes share similar high-level feature distributions, routing attractors may overlap across classes, potentially weakening class separation.
\section{Conclusions}
We introduced \textbf{Routing without Forgetting (RwF)}, a transformer-based approach that reframes continual learning as a routing problem rather than purely a parameter specialization problem. Instead of allocating or merging task-specific modules, RwF performs input-conditioned routing through energy-based associative retrieval, implemented via Hopfield Pooling layers embedded directly within the backbone.

Across multiple \emph{Online Class-IL} benchmarks, this architectural shift yields consistent improvements. On Split-ImageNet-R and Split-ImageNet-S, RwF surpasses strong prompt-, LoRA-, and dual-backbone baselines by clear margins, while remaining competitive on Split-CIFAR100. These gains are achieved with limited parameter growth and modest overhead, indicating that improvements arise from reshaping representation flow rather than increasing capacity. Ablations further show that routing depth modulates performance, with early-layer associative retrieval providing the best trade-off between accuracy and efficiency.

Architecturally, RwF introduces a many-to-few associative routing mechanism within transformer blocks. Unlike cross-attention with fixed learnable queries, RwF recomputes a small set of input-conditioned routing prompts at every forward pass through closed-form associative retrieval. Unlike prior Hopfield-based memory modules, retrieval operates inside the backbone to dynamically reallocate representational subspaces rather than storing external patterns or task-indexed prompts.

More broadly, our results suggest that stability in continual learning can emerge from architectural mechanisms that smoothly reorganize representation flow, rather than relying solely on gradient constraints, replay buffers, or explicit expert partitioning.

\bibliographystyle{unsrt}
\bibliography{references}  






\end{document}